\documentclass[letterpaper, 10 pt, conference]{ieeeconf}  

\IEEEoverridecommandlockouts                              

\usepackage{siunitx}
\usepackage{graphicx}
\usepackage{array}
\usepackage{url}
\usepackage{bm}
\usepackage{comment}
\usepackage{amsfonts}
\usepackage{psfig}
\usepackage{epsfig} 
\usepackage{mathptmx} 
\usepackage{times} 
\usepackage{amsmath} 
\usepackage{amssymb}  
\usepackage{multicol} 
\usepackage{upgreek}
\usepackage{bm} 
\usepackage{cite}
\usepackage{subeqnarray}
\usepackage{cases}
\usepackage{multirow}
\usepackage{epsfig} 
\usepackage{mathptmx} 
\usepackage{times} 
\usepackage{booktabs}
\usepackage{float}
\usepackage{sansmath}
\usepackage{subcaption}
\usepackage{color}
\usepackage{relsize}
\usepackage{textcomp}
\usepackage{svg}

\newtheorem{remark}{\rm\textbf{Remark}}
\usepackage{algorithm}
\usepackage{algpseudocode}

\title{\LARGE \bf
Fast Motion Planning for Non-Holonomic Mobile Robots via a Rectangular Corridor Representation of Structured Environments 

}
\author{
Alejandro Gonzalez-Garcia, Sebastiaan Wyns, Sonia De Santis, Jan Swevers and Wilm Decr\'e
\thanks{This work was supported by the Flanders Make SBO project ARENA (Agile \& Reliable Navigation). }
\thanks{Authors are with MECO Research Team, Department of Mechanical Engineering, KU Leuven, Belgium and with Flanders Make@KU Leuven, Belgium. {\tt\small \{alex.gonzalezgarcia, sonia.desantis, jan.swevers, wilm.decre\}@kuleuven.be}}
}

\begin{document}

\maketitle
\thispagestyle{empty}
\pagestyle{empty}

\begin{abstract}

We present a complete framework for fast motion planning of non-holonomic autonomous mobile robots in highly complex but structured environments. Conventional grid-based planners struggle with scalability, while many kinematically-feasible planners impose a significant computational burden due to their search space complexity. To overcome these limitations, our approach introduces a deterministic free-space decomposition that creates a compact graph of overlapping rectangular corridors. This method enables a significant reduction in the search space, without sacrificing path resolution. The framework then performs online motion planning by finding a sequence of rectangles and generating a near-time-optimal, kinematically-feasible trajectory using an analytical planner. The result is a highly efficient solution for large-scale navigation. We validate our framework through extensive simulations and on a physical robot. The implementation is publicly available as open-source software.

\end{abstract}

\section{Introduction}

\begin{figure*}[tb]
    \centering
    \includegraphics[width=0.8\linewidth] {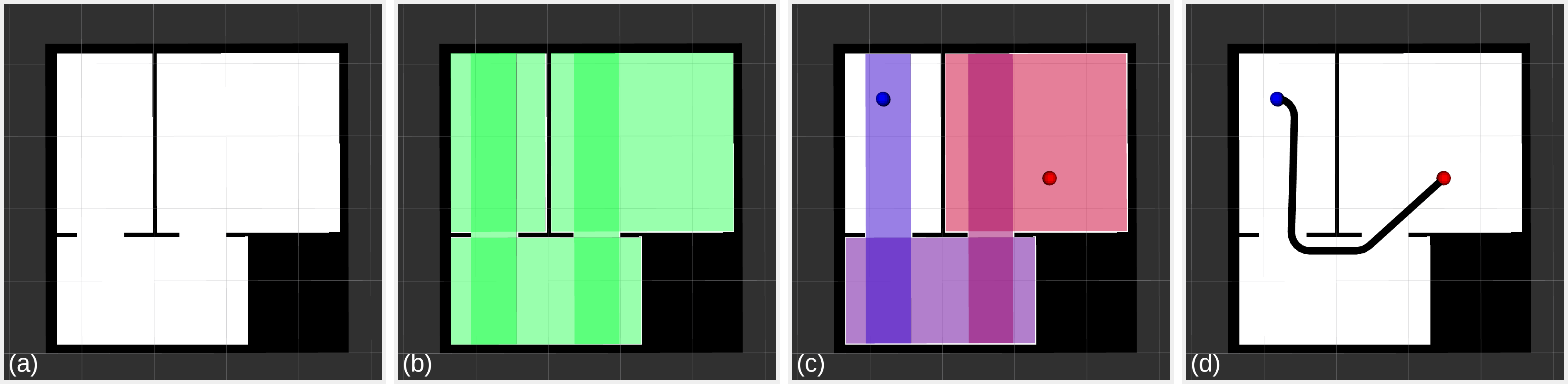}
    \caption{Example illustrating the corridor-based motion planning framework: (a) input occupancy grid, (b) corridor decomposition, (c) planned corridor sequence (blue to red), and (d) generated analytical trajectory.}
    \label{fig:framework}
\end{figure*}

Autonomous Mobile Robots (AMRs) are increasingly deployed in industries, such as manufacturing, warehousing, terminals, hospitals, smart farms, and greenhouses
\cite{FRAGAPANE2021405,DACOSTABARROS2021103729,9791350,young2019design,6907989}.
A central challenge in autonomy is motion planning, where a trajectory or path is computed from one position to another, balancing reliability, computational efficiency, and trajectory quality \cite{tzafestas2018mobile,Marcucci-RSS-25}. In this article, we focus on motion planning for non-holonomic AMRs operating in highly complex but structured environments, such as large factory floors or buildings with long corridors and narrow passages that can be particularly challenging for traditional planners.

Early motion planning methods, such as A* \cite{4082128} or Dijkstra \cite{dijkstra1959note}, rely on occupancy grid representations \cite{30720}. These methods can reliably provide collision-free paths, but as the map size or resolution increases, the number of grid cells grows rapidly, leading to longer planning times. Additionally, these paths are purely geometric, ignoring the non-holonomic constraints of AMRs. To address this limitation, lattice-based planners introduced precomputed motion primitives that enforce kinematic constraints during the search. These primitives can be generated offline either through closed-form solutions \cite{Pivtoraiko2009DifferentiallyLattices,doi:10.1177/0278364909340445} or optimization-based formulations \cite{Botros2023Spatio-TemporalPrimitives, Bergman2021ImprovedControl}. However, they still suffer from the scalability limitations of grid-based methods. While downsampling the map can mitigate these issues, it may also remove narrow doorways critical for successful navigation.

Sampling-based planners, such as Probabilistic Roadmaps (PRMs) \cite{508439} and Rapidly-Exploring Random Trees (RRTs) \cite{844730}, avoid exhaustively exploring the map. However, these probabilistic methods provide no deterministic guarantees, their solution quality strongly depends on the sampling density, and, as illustrated in \cite{6722915}, they often struggle in narrow passages, a common characteristic of structured environments with long hallways and door-like scenarios.

An alternative approach is planning through convex covers, by decomposing the environment into safe sets \cite{Marcucci2024,10935632,7138978,Liu2017PlanningEnvironments,10970076}. \cite{Marcucci2024} demonstrates the value of separating this decomposition into offline preprocessing and online planning phases, using precomputed safe boxes to accelerate motion planning. However, most of these methods focus on local decomposition around a path rather than the entire environment. These methods typically start with a collision-free path from a discrete planner, then inflate geometric shapes around this path to construct convex covers of the free space. This process generally begins by finding inscribed ellipses from points, lines, or polytopes \cite{10970076,Liu2017PlanningEnvironments,werner-RSS-25}, and converting them into polygonal obstacle-free regions through iterative optimization. Here, the coverage and predictability of the safe sets are dependent on the initial seed. Moreover, trajectories are commonly generated using mixed-integer programming \cite{7138978}, RRT* \cite{10970076}, or piecewise polynomial optimization \cite{Marcucci-RSS-25} within the convex covers, or through joint optimization of the trajectory and the convex cover \cite{10935632}. Nevertheless, most decomposition-based methods target systems with free motion in Cartesian space, such as quadrotors or point-mass models, rather than kinematically constrained systems. In contrast, \cite{SoniaECC} has shown that near-time-optimal trajectories for non-holonomic unicycle robots can be computed analytically through predefined sequences of rectangular corridors, i.e., rectangular regions that decompose the free space. These trajectories were validated against optimal control problem (OCP) solutions, showing a two-order-of-magnitude reduction in computation time while remaining mostly within $<1\%$ of the time-optimal solution. 

Despite recent advances, existing approaches still exhibit key limitations to achieve real-time motion planning for non-holonomic AMRs in structured but complex 2D environments. Many methods scale badly with map size and resolution, rely on heuristics or sampling, ignore kinematic constraints, or require expensive online computations. To address these challenges, we propose a framework that combines offline free-space decomposition to manage environmental complexity and online analytical planning, entirely avoiding online optimization. In the offline phase, given a map, the free space is decomposed into overlapping rectangles, hereafter referred to as corridors, a process only repeated if the map changes. This corridor-based representation yields a compact, search-based graph that is inherently collision-free. Unlike grid downsampling approaches, our method provides structural geometric compression, preserving all navigable passages. In the online phase, this precomputed decomposition is used to find a sequence of corridors and leverages analytical methods to generate near-time-optimal, kinematically feasible trajectories in real time. Unlike existing approaches, this framework enables real-time, optimization-free motion planning that scales with structural complexity rather than resolution, as illustrated in Fig.~\ref{fig:framework}.

\subsection{Contributions}

This paper introduces a complete framework for fast motion planning of non-holonomic AMRs in highly complex but structured environments. The main contributions of this work include:

\begin{itemize}
    \item A deterministic, compact representation: We propose a novel algorithm to deterministically decompose the entire free space into a compact, collision-free graph of rectangular corridors. This representation achieves structural compression ratios exceeding 10,000:1, and can be reused across different planning algorithms, not just our specific framework.
    \item Real-time, kinematically-feasible motion planning framework: Our approach computes a corridor sequence and generates near-time-optimal, kinematically feasible trajectories that include both geometry and timing. Compared to planners that produce only geometric paths, our method achieves planning times up to an order of magnitude faster, making it suitable for real-time operation in complex environments.
    \item Comprehensive experimental validation: The framework's performance and reliability are demonstrated extensively through simulation on multiple layout maps and on a real robot operating in a laboratory environment.
    \item Open-source implementation: The framework is implemented in ROS 2 and designed for reproducibility. The source code, and simulation environments are available at: \url{https://github.com/alexglzg/corridor_navigation}.
\end{itemize}

\section{Preliminaries}

In this section, we present the problem formulation and an overview of our motion planning framework.

\subsection{Problem Formulation}

We consider a non-holonomic AMR, common in industrial settings, modeled as a unicycle with state $\mathbf{x} = [x, y, \theta]^\top \in SE(2)$, where $(x,y)$ denotes the position and $\theta$ the orientation. The robot has a circular footprint of radius $a$. The control vector is $
\mathbf{u} = [v, \omega]^\top \in \mathcal{U}, 
\; \mathcal{U} = [0, v_{\max}] \times [-\omega_{\max}, \omega_{\max}]$,  with $v$ the translational velocity and $\omega$ the angular velocity. The kinematic model follows:
\begin{equation}\label{eq:motion_eq}
\dot{x} = v\cos\theta, \quad \dot{y} = v\sin\theta, \quad \dot{\theta} = \omega.
\end{equation}

The environment is represented as an occupancy grid $M \in \{0,1\}^{m \times n}$ with resolution $\delta$ meters per pixel, where $M_{ij} = 0$ denotes free space and $M_{ij} = 1$ denotes obstacles. The collision-free configuration space is defined as $\mathcal{C}_{\text{free}} = \{\mathbf{x} \in SE(2) : B(\mathbf{x}, a) \subset \mathcal{F}\}$, where $B(\mathbf{x}, a)$ represents the robot's footprint at pose $\mathbf{x}$ and $\mathcal{F}$ is the free space in the workspace. Given a start pose $\mathbf{x}_s$ and goal pose $\mathbf{x}_g$, the motion planning problem seeks a trajectory $\tau: [0,T] \rightarrow SE(2)$ that minimizes the traversal time $T$ subject to: (i) boundary conditions $\tau(0) = \mathbf{x}_s$ and $\tau(T) = \mathbf{x}_g$, (ii) collision avoidance $\tau(t) \in \mathcal{C}_{\text{free}}$ for all $t \in [0,T]$, and (iii) kinematic feasibility under the control constraints $\mathcal{U}$.

Traditional grid-based methods discretize this problem over $O(mn)$ cells, leading to computational complexity that scales with map resolution rather than environment complexity. This motivates our corridor-based decomposition, which reduces the search space to a compact graph whose size depends on the structural complexity of the environment. We define a rectangular corridor as a tuple 
\(r = (\mathbf{c}, \mathbf{d}, \phi)\), where 
\(\mathbf{c} \in \mathbb{R}^2\) is the center, 
\(\mathbf{d} = [w,h]^\top \in \mathbb{R}^2_{>0}\) specifies the width and height, 
and \(\phi \in [0,2\pi)\) is the orientation. 
The corridor corresponds to the region
\begin{equation}
\mathcal{A}(r) = \{ \mathbf{x} \in \mathbb{R}^2 : 
|R_\phi^\top (\mathbf{x} - \mathbf{c})| \leq \tfrac{1}{2}\mathbf{d} \},
\end{equation}
where \(R_\phi\) is the rotation matrix and the inequality is interpreted element-wise.

\subsection{Architecture Overview}

We briefly outline the proposed framework and its key elements, which will be explained in detail in the subsequent sections.
Our framework employs a two-phase approach that decouples environment representation from trajectory generation, as illustrated in Fig.~\ref{fig:framework}. In the offline phase, the occupancy grid undergoes corridor decomposition to produce a set of overlapping rectangles $\mathcal{R} = \{r_1, \dotsc, r_{n_r}\}$, where $n_r = |\mathcal{R}|$, and their connectivity graph $G = (\mathcal{R}, E)$, where $E$ encodes adjacency relations through rectangle overlaps. From this representation, we precompute a transition graph $G_T = (V_T, E_T)$, where $V_T$ are feasible points to enter, travel, and exit corridors, and $E_T$ the straight-line connections between them, with $|V_T| \ll mn$ for structured environments.

The online phase processes planning queries through three sequential steps: (i) augmenting $G_T$ with start and goal poses, (ii) finding the shortest path in $G_T$ using Dijkstra's algorithm to obtain a corridor sequence $\mathcal{S}$, and (iii) generating a near-time-optimal trajectory through $\mathcal{S}$ using analytical methods. This separation enables real-time performance with planning complexity $O(E_T + |V_T| \log |V_T|)$, independent of the map resolution.

\section{Automatic Corridor Generation}

\begin{figure*}[tb]
    \centering
    \includegraphics[width=0.8\linewidth] {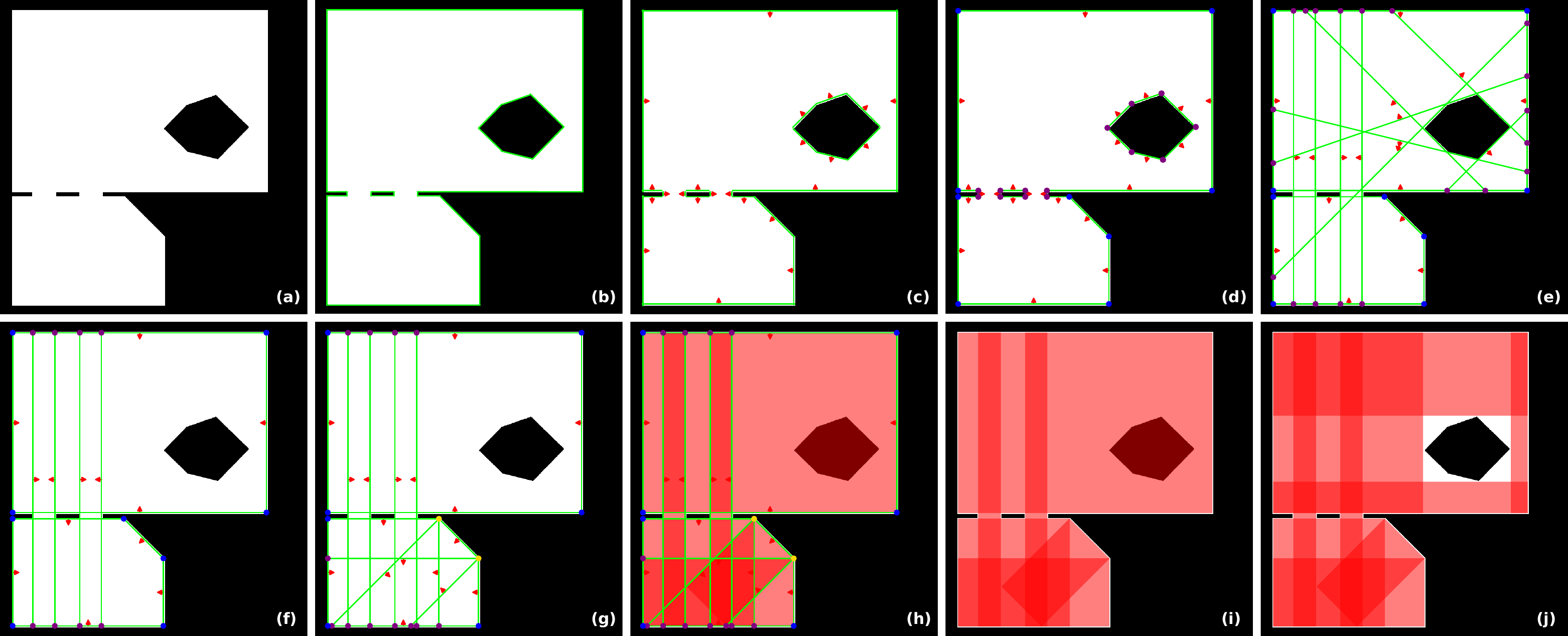}
    \caption{Automatic corridor generation pipeline example. (a) Floor plan as a binary image. (b) Detected line segments. (c) Straightened and shifted line segments. (d) Snap point extraction, blue dots mark full snap points, purple dots mark half snap points. (e) Closed rooms and continuous hallways after extending half snap points. (f) Pruned snap graph after obstacle removal. (g) Full snap extension, yellow dots mark the new double snap points. (h) Usage of snap points and faces to construct maximal axis-aligned rectangles. (i) Rectangle generation before obstacle carving. (j) Rectangles overlapping obstacles are split into fragments, resulting in the final coverage. }
    \label{fig:pipeline}
\end{figure*}

In this section, the Automatic Corridor Generation (ACG) algorithm is described. Fig.~\ref{fig:pipeline} illustrates the full corridor extraction pipeline.

\subsection{Design Objectives}

The proposed algorithm extracts corridors from 2D occupancy grids to create a compact spatial representation for efficient motion planning in structured indoor environments. Given an occupancy grid $M \in \{0,1\}^{m \times n}$, a set of rectangles $\mathcal{R}$ and their connectivity graph $G = (\mathcal{R}, E)$ are computed, designed with four key objectives:

\begin{enumerate}
    \item \textbf{Safety:} all corridors lie entirely within free space through explicit clearance margins;

    \item \textbf{Coverage:} maximize free-space coverage while maintaining geometric simplicity;

    \item \textbf{Compactness:} minimize the number of rectangles to reduce graph complexity;

    \item \textbf{Efficiency:} achieve polynomial-time complexity scaling with structural features rather than map resolution.
\end{enumerate}

\subsection{Algorithm}

\subsubsection{Stage 1-2, Line Detection and Straightening} 
We detect wall segments using a line segment detector and cluster them by orientation. Segments are aligned to canonical directions (e.g., $\mathcal{D} = \{\ang{0}, \ang{90}\}$) when within tolerance, or to their mean angle otherwise. For each segment, we compute its unit direction and project the original endpoints onto the line through their midpoint, yielding aligned endpoints $\mathbf{P}'$ and $\mathbf{Q}'$. To ensure safety, we determine the inward normal $\mathbf{n}$ by sampling the occupancy grid on both sides and selecting the direction with maximum free-space samples, such that $\mathbf{n}$ always points into navigable space. Each line is shifted inward by a clearance margin expressed in pixels, $\rho$, along the inward normal, giving a shifted line segment $\ell = \{\mathbf{P}' + \rho\mathbf{n}, \mathbf{Q}' + \rho\mathbf{n}\}$, which provides collision-free geometry.

\subsubsection{Stage 3-4, Snap Point Extraction and Extension} 
We cluster nearby endpoints within a threshold $d_s$ (snap distance, i.e., the maximum distance at which endpoints are merged) using union-find in $O(n_\ell\alpha(n_\ell))$ time, where $n_\ell$ is the number of line segments. Each cluster generates a snap point based on the incident line count and interior angle (see Fig.~\ref{fig:pipeline}(d)):
\begin{itemize}
    \item \emph{Full snap points} (convex corners, with angle of amplitude $< \ang{180}$ measured inside free space): placed at line intersection.
    \item \emph{Half snap points} (concave/hanging corners, with angle of amplitude $\geq \ang{180}$ measured inside free space): represented by two overlapping points linked as \emph{sisters}, so they can later extend in different directions to close gaps.
\end{itemize}
Half snap points (purple in Fig.~\ref{fig:pipeline}(d)) cast rays along their wall normal to find connection targets (represented by cast green rays in Fig.~\ref{fig:pipeline}(e)). Each ray finds the nearest valid wall intersection, after which the half snap either merges with a collinear counterpart or extends to the nearest wall hit, closing gaps in the corridor network.

\subsubsection{Stage 5-6, Face Identification and Corner Resolution}
We trace boundary cycles in the snap graph and classify each as a building interior (face=0) or an obstacle (face$>$0). For each cycle with center $\mathbf{c_c}$, we initialize a score $S=0$ and process each edge $i$ with length $l_i$, midpoint $\mathbf{m}_i$, and normal $\mathbf{n}_i$. We compute $d = \mathbf{n}_i \cdot (\mathbf{c_c} - \mathbf{m}_i)$: if $d > 0$, add $l_i$ to $S$; otherwise subtract $l_i$. Cycles with $S < 0$ are classified as obstacles. These snap points are grouped by connectivity and removed (see Fig.~\ref{fig:pipeline}(f)).

After obstacle removal, obtuse full snap points (with angle of amplitude $> \ang{90}$ measured inside free space) are converted to \emph{double snap points}. From each obtuse corner, we cast orthogonal rays along wall normals until they hit opposing walls, then insert extension lines to create \emph{double snap points} that decompose the obtuse angle into two $\ang{90}$ turns. This transformation ensures all corners are either $\ang{90}$ or can be decomposed into $\ang{90}$ turns (acute angles are retained as snap points, but do not contribute to rectangle generation), enabling axis-aligned rectangle generation. Double snap points (shown as yellow circles in Fig.~\ref{fig:pipeline}(g)) can spawn up to two rectangles from their orthogonal wall pairs.

\subsubsection{Stage 7-8, Rectangle Generation and Obstacle Carving}
We generate maximal axis-aligned rectangles by traversing snap points in priority order (double, full, then half). From each snap point, we follow incident walls to find potential rectangle corners. When four corners form a valid rectangle, it is added to the set. Half snap pairs with opposing normals define corridor rectangles connecting rooms. Successfully created rectangles then remove their corner snaps from working sets, preventing duplicates. 

When rectangle $r$ overlaps an obstacle, we split $r$ into up to four axis-aligned fragments that surround the obstacle's bounding box (see Fig.~\ref{fig:pipeline}(i)). Fragments with a non-positive area are discarded. This ensures safety while maintaining overlap connectivity around obstacles. Finally, if any rectangle does not comply with a minimum width or height to contain the robot footprint, it is discarded.

\subsubsection{Stage 9, Corridor Connectivity Graph Construction} \label{sec:cgc}
We construct the corridor connectivity graph $G = (\mathcal{R}, E)$ by testing all rectangle pairs for overlap using the Separating Axis Theorem (SAT). An edge $(r_i, r_j) \in E$ exists if rectangles overlap with sufficient area to contain the robot footprint. During overlap testing, we store the intersection polygon $\mathcal{I}_{ij} = \mathcal{A}(r_i) \cap \mathcal{A}(r_j)$ for each valid edge, as these geometries enable transition point extraction for motion planning (Section~\ref{sec:transition_graph}).

\begin{remark}
The corridor graph $G = (\mathcal{R}, E)$ provides a general spatial decomposition that can serve other planning algorithms, such as \cite{Marcucci2024}. While we construct a specific graph $G_T$ for point-to-point navigation, the corridor representation could be adapted for other structured environment tasks to leverage corridor areas or connectivity \cite{10610708}.
\end{remark}

\subsubsection{Complexity Analysis}
Dominant costs arise from line sorting $O(n_\ell\log n_\ell)$; snap clustering with union-find $O(n_\ell\alpha(n_\ell))$; snap extension requiring wall intersection tests $O(k_h \cdot n_\ell)$, where $k_h$ is the number of half snap points; rectangle generation with snap traversal $O(k^2)$ where $k$ is the number of snap points; and SAT-based overlap testing $O(n_r^2)$. Since $\alpha(n)$ is effectively constant for all practical values, and typically $k_h \leq k \approx n_\ell$ for structured maps, the corridor decomposition complexity simplifies to:
\begin{equation}
O(n_\ell\log n_\ell + kn_\ell + k^2 + n_r^2)
\end{equation}
where the rectangle count $n_r$ depends on map complexity rather than area, ensuring scalability.

\begin{remark}
    Within the proposed pipeline, the user can opt to include or ignore obstacles during corridor generation. This design choice, which may be based on environmental knowledge or the use of a local planner, allows the system to bypass the face identification (Fig.~\ref{fig:pipeline}(f)) and obstacle carving steps (Fig.~\ref{fig:pipeline}(j)), thereby reducing algorithmic complexity.
\end{remark}

\section{Corridor-Based Motion Planning}

In this section, we describe the pipeline for efficient planning based on the proposed rectangular corridor free-space representation.

\subsection{Transition Graph Construction}\label{sec:transition_graph}

Given the corridor connectivity graph $G = (\mathcal{R}, E)$ from Section~\ref{sec:cgc}, we construct a planning graph $G_T = (V_T, E_T)$ that transforms spatial relationships into a searchable structure. While $G$ captures which corridors connect, $G_T$ specifies where the robot can transition between them. The nodes $V_T$ consist of transition points, including corridor centers $\mathbf{c}_i$, and points extracted from the stored intersection geometries $\mathcal{I}_{ij}$, i.e., intersection centroids and corners. Edges connect points $\mathbf{p}_a, \mathbf{p}_b \in V_T$ if the line segment $\overline{\mathbf{p}_a\mathbf{p}_b}$ lies entirely within at least one corridor, ensuring collision-free paths. This precomputation executes once per map with complexity $O(|E| \cdot |V_T|^2)$.

\subsubsection{Corridor Sequence Planning}
Given start $\mathbf{p}_s$ and goal $\mathbf{p}_g$ positions, we identify their containing corridor and, if a direct path is viable within a single corridor, we use that path. Otherwise, we augment the precomputed $G_T$ with temporary nodes for the start and goal positions, connecting them to reachable transition points within their respective containing corridors. A shortest path is then computed using Dijkstra's algorithm, with edge weights $w(e)$ defined by a combination of Euclidean distance and a penalty for corridor transitions:
\begin{equation}
w(e) = \|\mathbf{p}_i - \mathbf{p}_j\|_2 + \lambda \cdot \mathbb{1}[\text{corridor transition}]
\end{equation}
where $\mathbf{p}_i, \mathbf{p}_j$ are the positions of the connected nodes, $\mathbb{1}[\cdot]$ is the indicator function and $\lambda \geq 0$ penalizes corridor changes. This shortest path yields a sequence of transition points $W = (\mathbf{p}_s, \mathbf{t}_1, \dotsc, \mathbf{t}_{n_t}, \mathbf{p}_g)$, where each $\mathbf{t}_i \in V_T$ represents a transition point between corridors. These transition points induce a corridor sequence $S = (s_1, \dotsc, s_{n_s}), \; n_s = |S|$, where each $s_j \in \mathcal{R}$, by tracking which corridors contain consecutive points. We remove redundant transitions to further optimize this sequence. Next, a traversal direction is computed for each corridor based on the waypoints, with angles snapped to the nearest axis-aligned direction. When a corridor exceeds a width/height or height/width set ratio, it follows its longest directed axis. The directed sequence $\tilde{S}= (\tilde{s}_1, \dotsc, \tilde{s}_{n_s})$ is then passed to a dedicated analytical planner (AP) that generates a smooth, near-time-optimal, and collision-free trajectory respecting the vehicle's dynamic constraints. The online sequence-planning process is dominated by the graph search, with a time complexity of $O(E_T + |V_T| \log |V_T|)$.

\begin{remark}
    The corridor sequence $S$ computation is a general solution independent of system dynamics or planning objectives. Thus, it can be paired with other algorithms for trajectory generation through convex sets, such as \cite{7138978,Marcucci-RSS-25}. 
\end{remark}

\subsection{Analytical Motion Planning}
The AP generates trajectories by concatenating time-optimal-based motion primitives within the free space defined by the directed corridor sequence $\tilde{S}$. The method presented in \cite{SoniaECC} delivered near time-optimal solutions in two-corridor scenarios, where OCP approaches were still tractable for comparison. Its slight suboptimality stems from heuristic rules used to place the time-optimal primitives in constrained environments. We extend this idea to sequences of two or more corridors, and consider additional heuristic rules to address maps containing long corridors and narrow passages. A brief description of the approach is provided below, with an emphasis on the new rules.

For the unicycle model (\ref{eq:motion_eq}), three time-optimal-based motion primitives are defined: on-the-spot rotations \(T^\bullet\) with \(v(t)=0\) and \(\omega(t)=\pm \omega_{\max}\), circular arcs \(C^\bullet\) with \(v(t)=v_{\max}\) and \(\omega(t)=\pm \omega_{\max}\) (turning radius \(\rho_t = v_{\max}/\omega_{\max}\)), and straight line segments \(S\) with \(v(t)=v_{\max}\) and \(\omega(t)=0\); in \(T^\bullet\) and \(C^\bullet\), \(\bullet \in \{+,-\}\) indicates the sign of \(\omega(t)\).


The key principle behind the planner is to decompose the trajectory computation into smaller, decoupled pieces. We obtain this subdivision by placing an \textit{intermediate circle} $o_j$ with radius $\rho_t$ between each two consecutive corridors \((\tilde{s}_j,\tilde{s}_{j+1})\) for \(j=1,\dotsc ,n_s-1\), where $n_s = |\tilde{S}|\geq 2$. This circle serves as an intermediate goal, guiding the robot from one corridor to the next while ensuring it remains within the corridor boundaries. We place the center of each $o_j$ either to the right or to the left of $(\tilde{s}_j, \tilde{s}_{j+1})$, relative to their traversal direction, depending on whether a clockwise or counterclockwise rotation is required to align $\tilde{s}_j$ with $\tilde{s}_{j+1}$. Accordingly, we perform the transition from $\tilde{s}_j$ to $\tilde{s}_{j+1}$ by executing a clockwise or counterclockwise circular arc along $o_j$. In contrast to \cite{SoniaECC}, a first additional rule is introduced for the case where \(\tilde{s}_j\) and \(\tilde{s}_{j+1}\) share the same traversal direction, which would make the placement of the intermediate circle \(o_j\) indeterminate. In this situation, we consider the relative orientation of \((\tilde{s}_j,\tilde{s}_{j+q})\), $q = 2$, and we repeat the procedure necessary by increasing $q$ until a change in direction is detected, or until the final corridor is reached. In the latter case, we determine the last corridor's rotation direction with the angle of the line connecting the center of the penultimate corridor to the final target position.

In general, the solution trajectory is composed of $2N + 3$ motion primitives: 
\begin{equation}\label{eq:general_traj}
T_1^\bullet C_2^{\bullet} S_3^{}C_4^{\bullet}S_5^{}C_6^{\bullet}\dotsc S_{2N-1}^{}C^\bullet_{2N}S_{2N+1}^{}C^\bullet_{2N+2}T^\bullet_{2N+3},
\end{equation}
where the subscripts indicate the order of appearance of each primitive in the sequence. The overall sequence is obtained by first computing independent trajectory pieces within each corridor, and then connecting them through the circular arcs $C_4^{\bullet}$, $C_6^{\bullet}$, $\dotsc$, $C_{2N}^{\bullet}$. The independent trajectory pieces are (i) the initial sequence $T_1^\bullet C_2^\bullet S_3^{}$, connecting the start pose $\mathbf{x}_s$ to $o_1$; (ii) the segments $S^{}_5$, $S^{}_7$, $\dotsc$, $S^{}_{2N-1}$, each of them connecting two consecutive circles $o_j, o_{j+1}, \; j = 1, \dotsc, n_s-2$; (iii) the final sequence $S^{}_{2N+1}C^\bullet_{2N+2}T^\bullet_{2N+3}$, connecting $o_{n_s-1}$ to the end pose $\mathbf{x}_g$. 

We introduce a second additional rule when two or more intermediate circles are closer than a distance $\rho_t$ and share the same direction of rotation. Such a situation often arises in door-like scenarios, where only a short portion of a corridor is traversed. In this case, we merge the circles into a single one, with its position adjusted to avoid collisions with the corridor walls. As a result, the number of motion primitives in the solution sequence (\ref{eq:general_traj}) is reduced. Finally, each pair of segments associated with an intermediate circle $o_j$ is checked for intersection. If the segments intersect, no arc maneuver is needed to move from $\tilde{s}_j$ to $\tilde{s}_{j+1}$, and the number of motion primitives in (\ref{eq:general_traj}) is reduced. In particular, the trajectory is updated depending on $j$: for $j=1$, the initial portion is recomputed to connect the start pose to $o_2$; for $j=2,\dotsc,n_s-2$, the two segments are replaced by a single segment connecting $o_{j-1}$ to $o_{j+1}$; and for $j=n_s-1$, the final portion is recomputed to connect $o_{n_s-1}$ to the goal pose. This process is repeated until no intersections remain.


\section{Results} \label{sec:results}

\begin{figure*}[tb]
    \centering
    \includegraphics[width=0.8\linewidth] {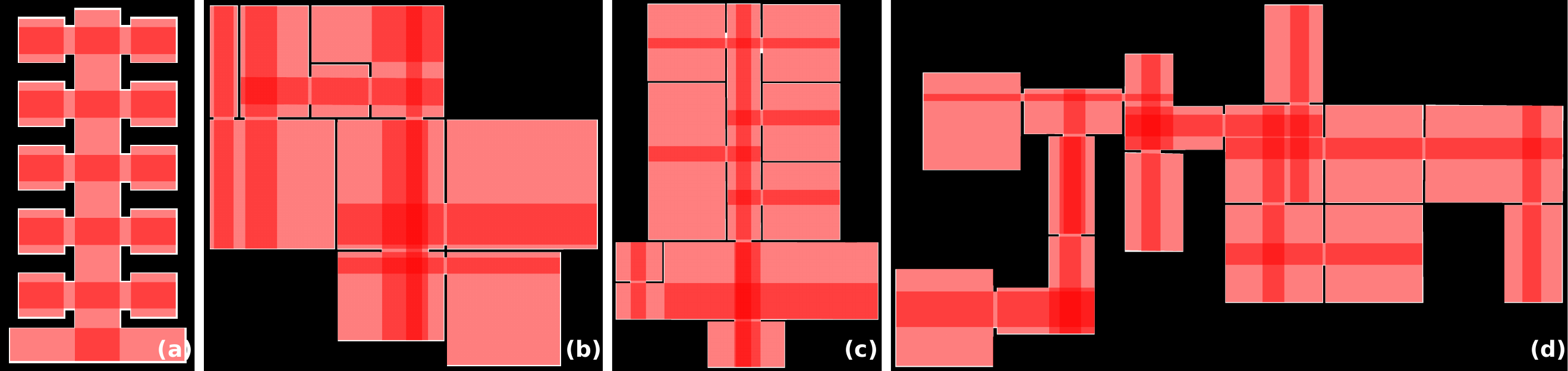}
    \caption{Representative Corridor Decompositions. (a) Small map (330x630 pixels) with 17 rectangles. (b)-(c) Large maps (1744x1624, 1738x2395 pixels) with 17 rectangles. (d) Large map (3444x1891 pixels) with 27 rectangles.}
    \label{fig:res_cg}
\end{figure*}

\begin{table*}
    \centering
    \caption{Corridor Generation Summary}
    \begin{tabular}{ccccccc}
        \toprule
        Category & $\#$ of Maps  & Avg Size (Pixels) & Rectangles $\in \mathcal{R}$ (Avg) & Nodes $\in V_T$ (Avg) & Avg Compression & Avg Time (ms) \\
        \midrule
        Small & 12 & 430K & 3-17 (10) & 15-117 (59) & 12,000:1 & 31 $\pm$ 15.90 \\
        Large & 12 & 3.9M & 11-27 (20) & 103-234 (166) & 24,000:1 & 169 $\pm$ 66.35 \\
        \midrule
        Overall & 24 & 2.2M & 3-27 (15) & 15-234 (113) & 18,000:1 & 100 $\pm$ 86.51 \\
         \bottomrule
    \end{tabular}
    \label{tab:corridorsummary}
\end{table*}

\begin{table}
    \centering
    \caption{Corridor Generation Scalability}
    \begin{tabular}{ccccc}
        \toprule
        Map Pixels & Nodes & Structural Compression & Time (ms) \\
        \midrule
        208K & 27 & 7,700:1 & 16\\
        1.3M & 103 & 12,500:1 & 67\\
        4.2M & 139 & 30,000:1 & 162\\
        7.3M & 159 & 46,000:1 & 279\\
         \bottomrule
    \end{tabular}
    \label{tab:corridorscalability}
\end{table}

\begin{table*}[tb]
\centering
\caption{Motion Planning Computational Performance Comparison}
\label{tab:plan}
\begin{tabular}{lcccccccc}
\toprule
Query Type & $\#$ of Samples & Path Length (m) & $\#$ of Corridors & \multicolumn{2}{c}{Proposed (ms)} & A* (ms) & H-A* (ms) & SLP (ms) \\
\cmidrule(lr){5-6}
&  &  &  & Framework & AP &  & & \\
\midrule
SHORT & 56 & 2-9 & 2-8 & 19.2 $\pm$ 9.2 & 1.28 $\pm$ 0.38 & 9.4 $\pm$ 6.9 & 101 $\pm$ 138.54 & 24 $\pm$ 15\\
MEDIUM & 32 & 10-24 & 3-11 & 22.2 $\pm$ 6.0 & 1.49 $\pm$ 0.32 & 25.5 $\pm$ 8.5 & 210 $\pm$ 212.44 & 126 $\pm$ 101 \\
LONG & 12 & 25-41 & 8-16 & 26.7 $\pm$ 4.1 & 1.77 $\pm$ 0.38 & 61.5 $\pm$ 14.0 & 530 $\pm$ 276.51 & 389 $\pm$ 137 \\
\midrule
OVERALL & 100 & 2-41 & 2-16 & 21.1 $\pm$ 8.1 & 1.41 $\pm$ 0.4 & 20.8 $\pm$ 18.8 & 176 $\pm$ 217.7 & 100 $\pm$ 138 \\
\bottomrule
\end{tabular}
\end{table*}

\begin{figure*}[tb]
    \centering
    \includegraphics[width=0.85\linewidth] {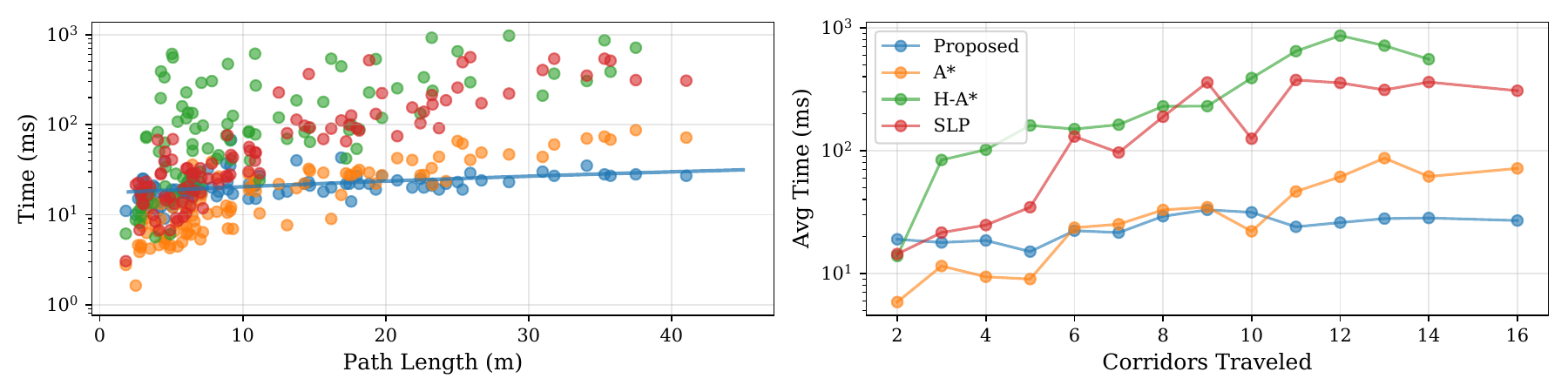}
    \caption{Motion Planning Benchmark. (Left) Computation Time vs Path Length. (Right) Average Computation Time vs Number of Corridors Traveled. Hybrid-A* failed to solve for some long paths. } 
    \label{fig:res_plan}
\end{figure*}

\begin{figure}[tb]
    \centering
    \includegraphics[width=0.7\linewidth] {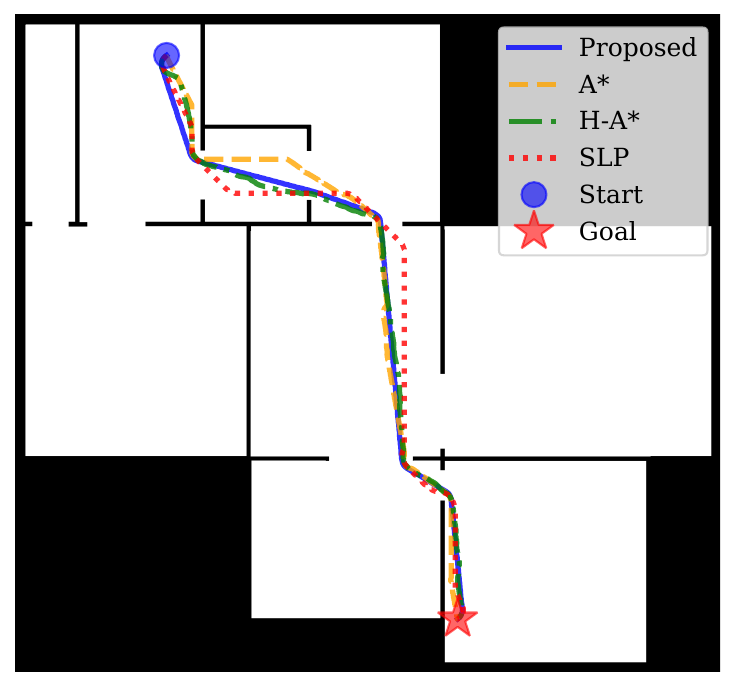}
    \caption{Illustrative Motion Planning Comparison. This medium query has a path of 18m, traveling through 8 corridors. Compared to A* (31.14ms), Smac Hybrid-A* (467.78ms), and Smac State Lattice Planner (119.28ms), the proposed approach (23ms) achieves the fastest computation time while planning a kinematically feasible trajectory.  } 
    \label{fig:map_plan}
\end{figure}

\begin{figure}[tb]
    \centering
    \includegraphics[width=0.85\linewidth] {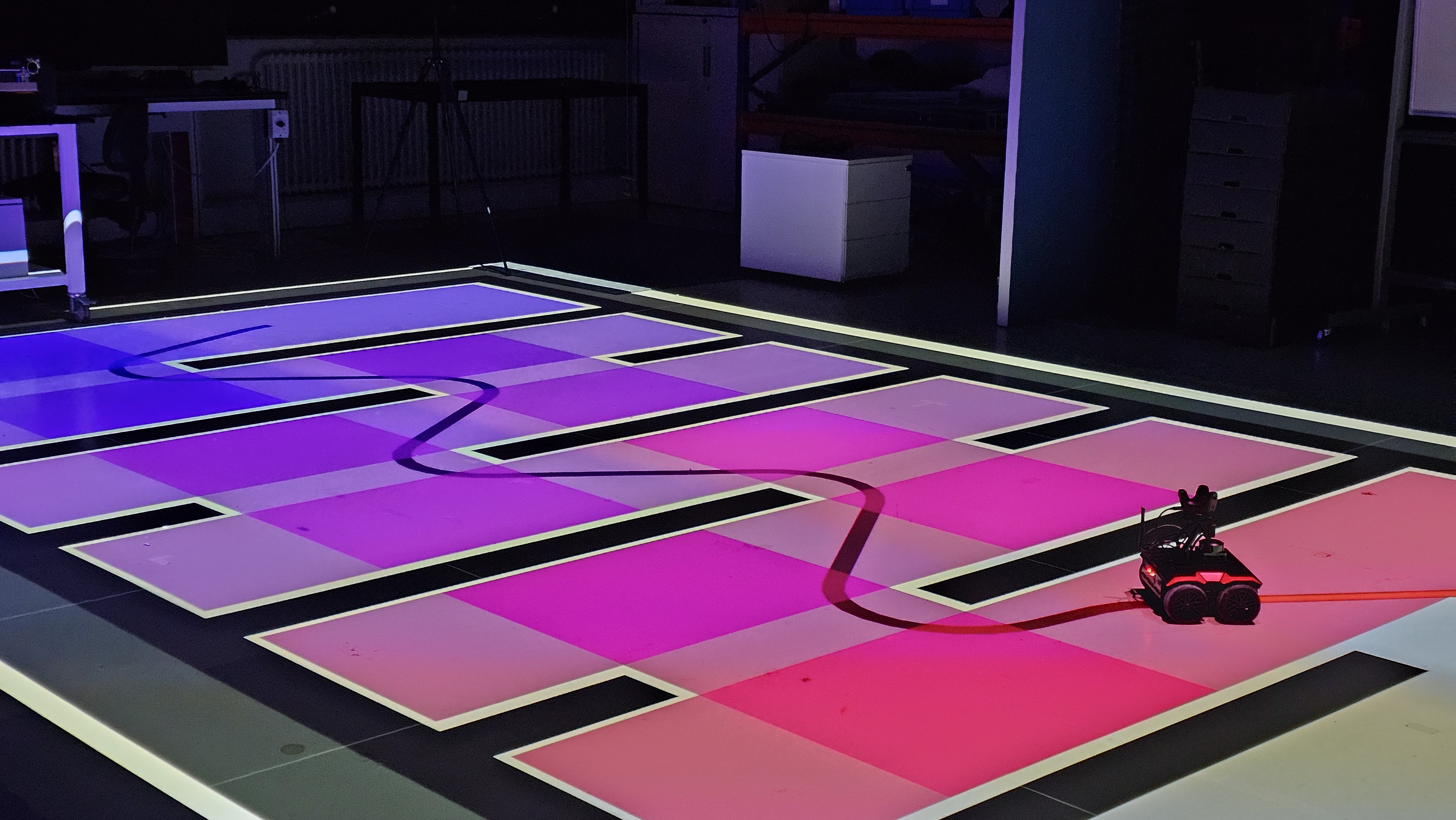}
    \caption{Experimental validation. Laboratory setup with ROSbot 3 and a projected virtual environment, showing a planned corridor sequence and trajectory pair.} 
    \label{fig:real}
\end{figure}

In this section, we present simulation and experimental results of our proposed motion planning framework. We begin by detailing the software implementation. Next, we provide a quantitative analysis of the proposed ACG algorithm. Then, we evaluate the performance of our complete motion planning framework against standard open-source planners from the Robot Operating System (ROS) 2 Nav2 Stack \cite{macenski2020marathon2}. Finally, we present an experimental validation with a physical robot.

\subsection{Software Implementation}

The proposed framework was implemented as a modular prototype in \texttt{Python} within the ROS 2 \cite{macenski2022robot} ecosystem. Graph operations are handled with \texttt{NetworkX} \cite{SciPyProceedings_11}. The architecture is composed of two primary ROS nodes, which separate the offline corridor generation process from the real-time online planning tasks. All simulations were conducted on an Intel Core i7-12800H CPU with 32 GB RAM running Ubuntu 24.04 LTS and ROS 2 Jazzy inside a Docker container.

\subsection{Automatic Corridor Generation}

We present an evaluation of the proposed ACG algorithm on 24 synthetic environments, ranging from 330x630 to 2430x3010 pixels, mostly drawn from a large-scale dataset of indoor layouts \cite{9341284}. While our method can handle environments with sparse obstacles, we evaluate on obstacle-free maps to focus on the algorithm's core strength: capturing global navigational structure. In practice, dense local obstacles are better addressed through hierarchical planning, where our corridor decomposition provides global routes and local planners can handle dynamic obstacles \cite{tzafestas2018mobile}. Table~\ref{tab:corridorsummary} summarizes the overall performance, demonstrating average compression ratios of 18,000:1 with generation times under 300ms for all tested maps. The algorithm shows consistent behavior across map categories, with the number of nodes scaling primarily with environmental complexity (3-27 rectangles) rather than resolution. Table~\ref{tab:corridorscalability} further illustrates this scalability through representative examples. This structural compression enables storing a 7.3M-pixel map ($\approx$7MB uncompressed) as a 159-node graph ($<$10KB), achieving over a 10,000$\times$ memory reduction critical for embedded AMR systems. Fig.~\ref{fig:res_cg} shows illustrative examples of the corridor generation. 

\subsection{Motion Planning Framework}

We performed a comparison of the proposed motion planning framework with open-source motion planners, taken from the ROS 2 Nav2 Stack. We compared against their A* implementation, which provides a baseline for computational performance without considering kinematic feasibility, and against their Smac Hybrid-A* (H-A*) and their Smac State Lattice Planner (SLP), which consider kinematic feasibility for differential drive AMRs through Reeds-Shepp curves and motion primitives, respectively \cite{macenski2024smac}. We considered a footprint radius of 0.34m, $v_{max}=0.5$m/s, $\omega_{max}=2.0$rad/s. For H-A* and SLP, we used the same turning radius $\rho_t=0.25$m, and 16 headings to build the motion primitives. We evaluated 10 maps, testing 10 queries of start-goal pose pairs per map. Through these maps, we tested with increasing path lengths and complexity, measured with the number of corridors traveled, to evaluate the planning time and scalability. A summary of the results is described in Table~\ref{tab:plan}. Through different configurations of map size, path length, and corridors traveled, our motion planning framework consistently achieves planning times of 20-40ms. Compared to A*, our proposed framework achieves competitive computational performance on paths shorter than 10m, while achieving a speedup of 2.3$\times$ on paths larger than 25m, despite the \texttt{Python} interpreted language overhead. This suggests that a high-performance C++ implementation could achieve even faster planning times. In comparison to the kinematically-feasible planners, we achieve computation times an order of magnitude faster on medium and long paths while computing trajectories instead of geometric paths. In addition, most of the overhead in our method comes with the graph augmentation with start/goal points, and the graph search, as the AP remains consistently around 1-2ms, regardless of the path length or corridors traveled. Fig.~\ref{fig:res_plan} further illustrates the scalability properties of our method in comparison to A*, H-A*, and SLP. Fig.~\ref{fig:map_plan} shows an example query, where our proposed approach achieves the fastest planning time.

\subsection{Experimental Evaluation}

The proposed motion planning framework was deployed using a differential drive four-wheeled mobile platform, the Husarion ROSbot 3, with an onboard Raspberry Pi 5 running ROS 2. The robot is localized using an HTC VIVE Tracker 3.0 mounted on the vehicle, along with three HTC VIVE Base Stations. Virtual environmental features are visualized on the laboratory floor using four projectors mounted on the ceiling. With this setup, it is possible to project occupancy grid maps, current corridor sequences, motion plans, etc. In addition to the motion planning ROS nodes, a low-level controller node based on Model Predictive Control (MPC), and a localization node were deployed. The MPC was formulated using \texttt{CasADi} \cite{Andersson2019} and trajectory optimization solver \texttt{FATROP} \cite{vanroye2023fatrop}. Sensing and actuation ran onboard the robot, whereas the MPC and motion planning framework nodes ran off-board on an Intel Core i9-9900X CPU $@$ 3.50GHz with 16 GB RAM. Fig.~\ref{fig:real} shows the experimental setup. We validated the framework across 8 different map configurations of 3×6m, with planning times averaging 20$\pm$5ms,  and corridor generation under 40ms, consistent with simulation results. The accompanying video demonstrates real-time map switching, planning, and replanning.

\section{Conclusion}

This paper introduced a framework for motion planning of non-holonomic AMRs based on a deterministic rectangular corridor free-space decomposition. By reducing the search space to a graph whose size depends on map complexity rather than resolution, exceeding structural compression ratios of 10,000:1, the method achieves a 2.3$\times$ speedup on long paths compared to conventional grid-based approaches, while directly generating near-time-optimal, kinematically feasible trajectories. In comparison to kinematically-feasible geometric path planners, our framework shows computing times an order of magnitude faster. The framework's efficiency was validated in simulation across a variety of maps and experimentally on a real robot. 

Despite these advantages, the framework has some limitations. The corridor construction method is tailored for structured maps and may face challenges in environments with irregular geometry or fine features. While our method excels in scenarios where a significant reduction in search space is achieved, its benefits diminish on short paths or in few-corridor cases. Moreover, the AP currently enforces only velocity constraints, leaving acceleration limits to be addressed by local planning/control layers. Additionally, the AP's heuristic rules are tailored for expected cases, so unforeseen situations continue to pose a challenge in practice.

Future work will focus on extending the corridor representation to more complex and cluttered environments (which are currently out of scope), developing tuning guidelines based on map resolution, extending the motion planning framework to axis-unaligned corridors, and incorporating bicycle-model vehicles.

\bibliographystyle{IEEEtran}

\bibliography{References}
\end{document}